# SyROCCo: Enhancing Systematic Reviews using Machine Learning

Zheng Fang[1], Miguel Arana-Catania[2], Felix-Anselm van Lier[3], Juliana Outes Velarde[3], Harry Bregazzi[3], Mara Airoldi[3], Eleanor Carter[3], Rob Procter[1*]

*Affiliations:*
[1] Department of Computer Science, University of Warwick and Alan Turing Institute for Data Science and AI, Coventry, United Kingdom
[2] School of Aerospace, Transport and Manufacturing, Cranfield University, United Kingdom
[3] Government Outcomes Lab, Blavatnik School of Government, University of Oxford, Oxford, United Kingdom
[*] *Corresponding author:* rob.procter@warwick.ac.uk

*Email address:*
Zheng Fang: Z.Fang.4@warwick.ac.uk
Miguel Arana-Catania: miguel.aranacatania@cranfield.ac.uk
Felix-Anselm van Lier: felix-anselm.vanlier@bsg.ox.ac.uk
Juliana Outes Velarde: juliana.outesvelarde@bsg.ox.ac.uk
Harry Bregazzi: harry.bregazzi@bsg.ox.ac.uk
Mara Airoldi: mara.airoldi@bsg.ox.ac.uk
Eleanor Carter: eleanor.carter@bsg.ox.ac.uk
Rob Procter: rob.procter@warwick.ac.uk

## ABSTRACT

The sheer number of research outputs published every year makes systematic reviewing increasingly time- and resource-intensive. This paper explores the use of machine learning techniques to help navigate the systematic review process. Machine learning has previously been used to reliably 'screen' articles for review – that is, identify relevant articles based on reviewers' inclusion criteria. The application of machine learning techniques to subsequent stages of a review, however, such as data extraction and evidence mapping, is in its infancy. We therefore set out to develop a series of tools that would assist in the profiling and analysis of 1,952 publications on the theme of 'outcomes-based contracting'. Tools were developed for the following tasks: assign publications into 'policy area' categories; identify and extract key information for evidence mapping, such as organisations, laws, and geographical information; connect the evidence base to an existing dataset on the same topic; and identify subgroups of articles that may share thematic content. An interactive tool using these techniques and a public dataset with their outputs have been released. Our results demonstrate the utility of machine learning techniques to enhance evidence accessibility and analysis within the systematic review processes. These efforts show promise in potentially yielding substantial efficiencies for future systematic reviewing and for broadening their analytical scope. Beyond this, our work suggests that there may be implications for the ease with which policymakers and practitioners can access evidence. While machine learning techniques seem poised to play a significant role in bridging the gap between research and policy by offering innovative ways of gathering, accessing, and analysing data from systematic reviews, we also highlight their current limitations and the need to exercise caution in their application, particularly given the potential for errors and biases.

**Policy Significance Statement**

Systematic reviews, despite aiming to bolster evidence-based policy, often face challenges in linking research to practice due to their time-intensive nature and narrow scope. Using the example of a large-scale systematic review consisting of highly heterogeneous and complex social science literature, this research highlights the potential of machine learning tools to speed up the efficacy of systematic reviews, and to broaden their analytical scope by offering a "birds-eye view" of a large set of publications, allowing for the identification of patterns, themes, and trends in a particular research field. It introduces a novel dashboard presentation of academic evidence, illustrating a means by which policymakers could transition from passive recipients to active explorers of the extracted information.

# 1. Introduction

A systematic review pools information from a set of studies on a clearly stated topic, using a replicable method to identify, describe, appraise, and synthesise research (Gough et al., 2017). Rigorous and transparent, systematic reviews can reliably establish what is known about a topic, and reveal gaps that require further study (Gough et al., 2017). With an exponential increase in the number of articles published in the past 20 years (Fire & Guestrin, 2019), however, it has become increasingly difficult for researchers to systematically review all of the publications that are relevant to a specific topic. Considerable time and resources are required to collect relevant papers and identify information pertinent to the research question at hand. A prior study indicated that the average time between registration and publication for reviews registered on PROSPERO (International Prospective Register of Systematic Reviews) is around 67 weeks (Borah et al., 2017). A more recent study estimated that a systematic review of around 10,000 articles would need at least 40 weeks (Bannach-Brown et al., 2019). Such a lengthy process can mean that review data are already out of date by the time of publication, and so might risk being misleading (Brock, 2019; Cumpston & Chandler, 2022; Shojania et al., 2007). The viability of existing systematic review processes is therefore questionable, which motivates exploring the use of machine learning techniques to speed up and support the process.

Machine learning tools have the potential to surface information from large volumes of text, which would otherwise be time-consuming or impossible to extract manually. Many studies have used machine learning tools to help with systematic reviews, reducing the human and financial resources required and the time needed to conduct the research (Bannach-Brown et al., 2019; Marshal & Wallace, 2019; Marshal & Brereton, 2015; Howard et al., 2016; Tsafnat et al., 2014; O'Mara-Eves, et al., 2015). Machine learning tools have been shown to be particularly effective in the screening phase of the systematic review process, i.e. to determine the likely relevance of articles to the research question by categorising them as 'relevant' or 'irrelevant'. The application of machine learning tools to subsequent phases of the systematic review process, however, is still in its infancy. It is therefore beneficial to identify innovative ways of using machine learning tools in all the phases of the systematic review process.

This paper introduces and provides an account of an initiative to utilise and further investigate novel applications of machine learning to assist in systematic review processes and expand their analytical capacity both for research and practice. The project team consisted of members of the Government Outcomes Lab (GO Lab), Blavatnik School of Government, University of Oxford, who possess domain expertise in systematic reviewing and cross-sector partnerships, and the natural language processing (NLP) group in the Department of Computer Science, University of Warwick. Together, we explored how state-of-the-art machine learning tools could be applied to accelerate the GO Lab's review of outcomes-based contracting (OBC). We therefore named the initiative the Systematic Review of Outcomes Contracts - Collaboration (SyROCCo). The objective was to develop a machine learning tool to help researchers, policymakers, and practitioners working on OBC to understand the state of the evidence more clearly, and to identify the studies of most relevance to their requirements. More details of the data used during the project and the topic involved in the review are given in sections 2 and 4.

To support the ambitious, policy-relevant objectives of the OBC review, the project team developed a series of supervised machine learning tools to facilitate different tasks within the systematic review. To summarise:
1. a preprocessing tool was developed to extract from the original full articles the specific parts of the content to be used in the posterior steps;
2. a text classification tool was used to sort studies into different policy area categories;
3. named entity recognition was used to identify and extract key organisations, laws, and geographical information associated with the outcomes contracts described in the studies;
4. the systematic review data was connected to one of the GO Lab's existing datasets to use their joint information in the analysis;
5. semantic text similarity was used to identify subgroups of studies which may share thematic content.

The outputs of these different tools were then combined and visualised in an interactive online dashboard, designed to maximise utility to the GO Lab's policy, practice, and research audiences.

As an additional contribution, we have publicly released the SyROCCo dataset produced by the project,[1] including the articles used in the review and all the results of the machine learning techniques applied to them as described in this article. The details of this new dataset are in Appendix B.

## 2. Research context: Systematic Review of Outcomes-Based Contracting

The machine learning tools described in this paper were developed and tested on a systematic review conducted by the GO Lab and Ecorys (Picker et al., 2021). This section's brief introduction of the review contextualises subsequent discussion of how we used machine learning tools to assist the process.

---
[1] SyROCCo dataset. https://zenodo.org/doi/10.5281/zenodo.12204303

The focus of the systematic review is 'outcomes-based contracting' (OBC). OBC is a model for the provision of public services wherein a service provider receives payment, in-part or in-full, upon the achievement of pre-agreed outcomes. There are multiple forms of OBC. It has been implemented in numerous countries, and applied to a range of policy areas, from education to environmental management (Government Outcomes Lab, n.d.). As such, the empirical evidence on OBC is fragmented, dispersed, and difficult to navigate. It includes a wide variety of study designs and content. The purpose of GO Lab's systematic review is to gather and curate all of the existing evidence on OBC. We aim to map the current state of the evidence, synthesise key findings from across the published studies, and provide accessible insights to our policymaker and practitioner audiences.

The review's objective therefore entails dealing with a large body of published material. The initial bibliographic search returned over 11,000 results. After manual screening to establish their potential relevance to our topic, ~2,000 studies remained.[2] To manually extract details on a uniform set of variables from across this large and heterogeneous set of publications is so labour-intensive as to be a practical impossibility. One way around this difficulty is to further filter the studies according to narrower inclusion criteria until the number of included texts becomes more manageable. Doing so leads to detailed enquiries into specific sub-themes, and this is indeed how the GO Lab and Ecorys managed early policy reports produced from the review (Bregazzi et al., 2022; Elsby et al., 2022). While this produces insights for specific topics and policy areas, the filtering process is time consuming, and it leaves much potentially useful information in the larger body of published material unaccounted for and inaccessible to a broader audience.

With the assistance of machine learning, however, meaningfully processing the content of 2,000 texts becomes a more realistic possibility. Large-scale categorisation and data extraction could reveal patterns, connections, and trends in a particular research area that would otherwise have gone unrecognised. It also offers the potential to support GO Lab's aim of engaging stakeholders beyond academia. We wanted to explore how machine learning might make a large evidence base accessible to relevant stakeholders, such as policymakers, and allow them to navigate it for policy-relevant insights. Finally, machine learning might allow us to connect the data from the systematic review to GO Lab's already existing 'INDIGO Impact Bond Dataset'.[3] Connecting both sources of information could advance the level of evidence readily available to our audience, linking INDIGO's record of specific projects to published studies of those projects.

In summary, we set out to test whether and how machine learning could assist multiple aspects of our systematic review, from data extraction to the presentation of evidence to OBC stakeholders.

---

[2] The collaboration with Warwick Universities' data scientists, which could have introduced more efficient automated methods, was established subsequent to this initial manual screening phase. However, the availability of a manually screened body (1) ensures a comprehensive initial data set for subsequent in-depth analysis, both using traditional methods and ML; (2) equips us with a robust training dataset for such nuanced classification tasks which are to be developed in the future.

[3] INDIGO is the International Network for Data on Impact and Government Outcomes. 'Impact bonds' are a specific form of OBC. The Impact Bond Dataset collects data on impact bond projects from all around the world.

The following two sections explain which tools we used, and how we applied them to our review.

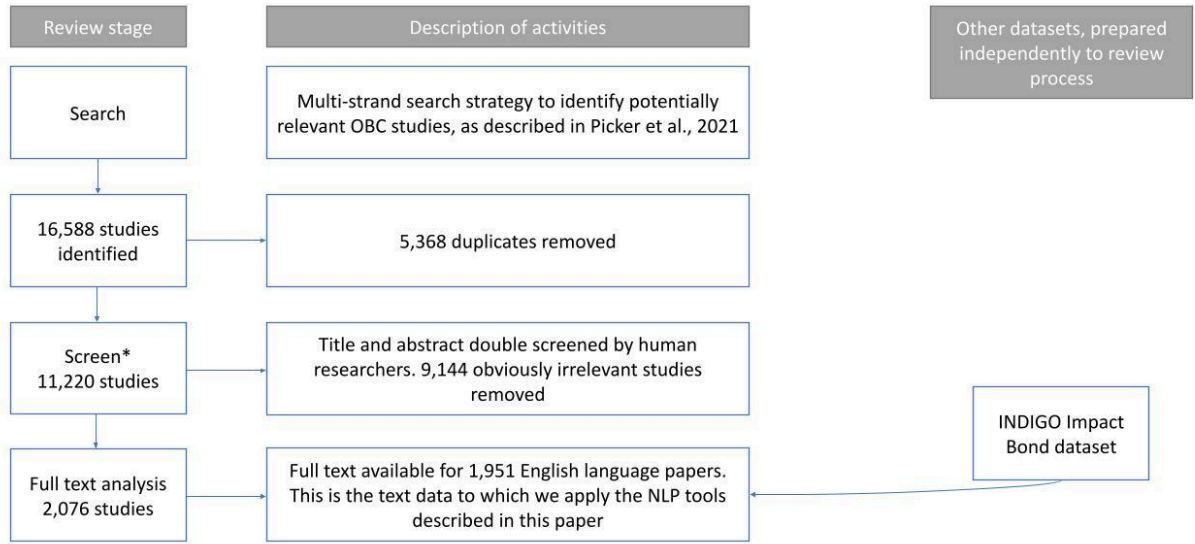

Figure 1: Description of the GO Lab systematic review process and application of NLP methods. The symbol '*' indicates the point at which ML tools are conventionally applied within the systematic review process.

## 3. Overview of Machine Learning Techniques

Several machine learning techniques were explored in this project, including text classification, named entity recognition, and semantic text similarity. An overview of these techniques is provided in this section.

**Text classification** is a machine learning technique designed to automatically classify unstructured text into predefined categories. Multiple previous studies have applied text classification in the screening phase of a review (Bannach-Brown et al., 2019; Marshal & Wallace, 2019; Marshal & Brereton, 2015; Howard et al., 2016; Tsafnat et al., 2014; O'Mara-Eves, et al., 2015). By automatically classifying articles as relevant or irrelevant to a particular research topic, such applications increase the efficiency of screening. However, machine learning techniques have shown to be effective and offer versatility for multiple classification types (Porciello et al. 2020). Classifying texts according to topics or geography, for example, can provide a rapid 'snapshot' of variety and distribution within an evidence-base (see Callaghan et al. 2021). Text classification can therefore also support the evidence-mapping stage of a systematic review.

To train a text classification model, conventional supervised approaches require a training set of open-ended texts that have been manually labelled into predefined categories, so the model can learn from how human reviewers categorise texts. The model's performance is then tested with a validation set, which also needs to be pre-labelled. Several strategies have been proposed to improve the accuracy of text classification models in systematic reviews (Wallace et al., 2010; Liu et al., 2018; Kontonatsios et al., 2017; Wallace et al., 2012; Miwa et al., 2014), all of which focus on screening more records to supplement the training set. In this study, instead of using a training set to train a text classification model, we used a semi-supervised approach, asking researchers to provide keywords describing each category. By learning from the keywords, the model is able to predict the probability of articles belonging to each category, and we can then ask human experts to determine a probability threshold for each category.

Compared with conventional approaches, the semi-supervised approach is more resource friendly. In particular, there is a significant decrease in the amount of human effort required to evaluate, comprehend and label a training data set. The approach followed is also more efficient for developing techniques to be applied after the initial screening phase. Other techniques are focused on filtering articles into just two broad categories (relevant/irrelevant), while in our approach the objective is to classify articles into many different categories according to the different aspects of the information that we want to explore. For instance, one classification may be according to policy domains, while another alternative classification may be done according to geographical areas.

**Named entity recognition (NER)** is a sub-task of information extraction that aims to identify different types of named entities mentioned in open-ended text such as companies, countries, laws, people, nationalities, etc. This information could correspond to a particular variable of interest (e.g., identifying key organisations located in a specific area). By automatically identifying these named entities, the time cost for researchers to locate variables of interest in large volumes of text is reduced. This can be further used to improve the screening efficiency by allowing researchers to select all articles that mention a specific named entity (such as a country). This could also give researchers a better idea of what kind of issues they are dealing with in the selected articles, and facilitates the organisation of articles into 'evidence maps'. In recent years, the use of pre-trained language models for named entity recognition has become increasingly attractive. Pre-trained named entity recognition models are large neural networks that are first pre-trained with large general corpora and then fine-tuned in a named entity recognition task. There are several advantages of using pre-trained named entity recognition models (Wang et al., 2020). First, it is easy to incorporate as it is not necessary to train a model from scratch. Second, it does not require labelled training data, which is important when dealing with a large number of texts. In this study, pre-trained named entity recognition models provided by the python packages 'spaCy'[4] and 'Flair'[5] (Akbik et al., 2019) were applied.

**Semantic text similarity** is a task in the field of natural language processing (NLP) that allows rating numerically how similar two texts are from a semantic point of view. Applied to a research article, it can be used to automatically identify further articles that talk about related subjects. It is useful for identifying articles in a particular subject area - perhaps revealing not-so-obvious connections between different texts that deserve further investigation. There are a number of approaches to computing the semantic similarity of texts, including the conventional 'Term Frequency-Inverse Document Frequency' (TF*IDF) (Salton et al., 1988) and more recently developed techniques based on pre-trained language models (Reimers & Gurevych, 2019; Yang et al., 2019; Devlin et al., 2018). Given a set of texts, these first encode the texts into a numerical representation that can be represented as a vector in an abstract space in which the directions each vector points to are related to the lexical semantics of the texts. Then these vector representations can be used to compute their semantic similarity by evaluating the distance between them. The TF*IDF matrix counts the occurrences of each word in each document, and weights terms according to how frequently they occur in the corpus (with infrequent terms weighted higher). Several semantic distance metrics exist, such as cosine similarity, Jensen-Shannon divergence and the dice coefficient (Mohammad & Hirst, 2012). The cosine similarity measures whether documents contain the same words with the same frequencies (weighted according to their corpus frequency). In this study TF*IDF method and the cosine similarity metric are used as they are simple to apply and memory efficient compared to other methods, which is important when dealing with a large number of texts and designing a production system to be used online. For future versions of this tool we will consider the implementation of other more complex similarity measures, and their evaluation both in terms of computational resources and results obtained.

---

[4] https://spacy.io/
[5] https://github.com/flairNLP/flair

## 4. Methodology

The project followed a co-production approach, with the GO Lab team and NLP group working together throughout. Having users at the heart of the development process has long been recognised as the key to successful IT projects (Bodker et al., 1975). This has become especially important as machine learning techniques, that are still unfamiliar to many, are now being widely applied to new products and services (Slota, 2020; Wolf, 2020). Objectives and progress were reviewed in weekly project team meetings. Initially, these focused on building common ground: (a) familiarising GO Lab team members with NLP and machine learning concepts, techniques, capabilities and limitations; (b) enabling the technical team members to gain an understanding of the systematic review process; and (c) establishing some initial requirements for how NLP tools could support it. As the project progressed, a series of prototypes became available for comments and feedback, enabling the requirements to be refined progressively.

**4.1 Data**

The data used to conduct the review consists of 1,952 individual publications on OBC. They include peer reviewed journal articles, book chapters, doctoral dissertations, and assorted 'grey literature'. The search and screening strategy for the identification of relevant studies is set out in the review protocol (Picker et al., 2021). The initial search in multiple bibliographic databases returned over 11,000 results (Figure 1). The research team screened the title and abstract of each result, excluding those that were obviously irrelevant to the topic of outcomes-based contracting, and keeping those that appeared relevant. This process narrowed down the body of texts to the final 1,952 that provide the content for the machine learning tool.

**4.2 Text extraction and pre-processing**

The articles meeting the inclusion criteria were collected in PDF format. To prepare the articles for application of machine learning tools, we extracted the texts from the PDF files and pre-processed them. We used the python package 'pdfminer'[6] to extract the text from the PDF articles. The pre-processing involved the following steps. First, we read the text in the UTF-8 format, normalised it using the library Unicodedata with the NFD decomposition, converted it to ASCII to simplify it, and then converted it back to UTF-8; this first step ensured a homogeneous version of all texts and eliminated unusual characters or codifications that could make the data analysis difficult. The second step involved removing the acknowledgements and bibliographic references from the texts. This information could confuse the data analysis, e.g. author names in the references could be extracted by the named entity recognition tool and get mixed up with the people mentioned in the text relevant to the subject of the article, references' publication years could be detected as dates relevant to the texts, etc. The deletion of references and acknowledgements was done using a rule-based approach based on the common formats of such sections, and refined iteratively based on its results. Third, we checked the processed articles against the original list of those screened by the research team (described above), to ensure the dataset was complete and matched the original.

---

[6] https://github.com/pdfminer/pdfminer.six

### 4.3 Connecting articles with the INDIGO Impact Bond Dataset

After obtaining the pre-processed and cleaned texts, we searched projects and organisations mentioned in the texts by using the names of projects and organisations provided by the Impact Bond Dataset. The Impact Bond Dataset had 250 projects and 1,094 organisations at the time of building SyROCCo (November 2021). The variables 'Name of project' and 'Alternative name(s) of project' were used to ensure that all of the studies in the evidence base that mention an INDIGO project could be identified. With regards to organisations, the Impact Bond Dataset also provided a list of organisations working on impact bond projects. However, some projects or organisations have short names that appeared in the texts as a generic word rather than as an organisational entity. For example, one impact bond project is called 'Aspire', which could be confused with the verb 'to aspire'. Overall, we identified 21 projects with names that could be confused for verbs or locations. To reduce the negative impact of this ambiguity, all suspicious project and organisation names in the dataset were manually identified, and then regular expressions were designed to extract text fragments that mentioned these names and these text fragments were checked by the GO Lab team. Based on the feedback, new search rules for suspicious names were devised.

Connecting the dataset of articles with the INDIGO dataset allows not only the current analysis to be complemented with the information from the INDIGO initiative, but also the Impact Bond Dataset to be expanded with information extracted in the current project. Connecting independent datasets adds value to the INDIGO initiative as it allows the production of new and more complex analyses that would be unfeasible otherwise. In addition, practitioners and researchers have expressed an interest in accessing the latest evidence in one place. Connecting these datasets helps INDIGO to be developed as a single comprehensive and user-oriented knowledge hub to help practitioners, policymakers and evaluators access the latest data and evidence on outcomes-based projects.

### 4.4 Detecting financial mechanisms, Sustainable Development Goals, and income levels

In order to improve the navigability of the results for researcher and policy audiences, mentions in the articles to other types of information were extracted: 1) financial mechanisms (type of outcomes-based instrument); the GO Lab team provided a list of terms related to those financial mechanisms based on prior discussions with a policy advisory group (Picker et al., 2021) and conceptual alignment; 2) references to the 17 Sustainable Development Goals (SDGs) defined by the United Nations General Assembly in the 2030 Agenda; 3) country names mentioned in each article and income levels related to the countries; world country codes[7] were used and the income level of each article determined by the income level of the countries mentioned in the article; for example, if an article mentions 'USA' and 'India' then it has an income level of 'high income' and 'middle income' as 'USA' and 'India' are respectively high and middle income countries according to the World Classification of Income Levels 2022 by the World Bank.

In all these cases a direct search of the relevant terms was performed, including regular expressions to improve the search results.

### 4.5 Policy areas identification

A text classification technique was applied to identify the corresponding policy areas of each article. The policy areas were pre-defined in seven categories: health, education, homelessness, criminal justice, employment and training, child and family welfare, and agriculture and environment. A query-driven topic model (QDTM) (Fang et al., 2021) was used to determine the probability of an article belonging to each policy area, using all text of the article as input. The QDTM is a semi-supervised machine learning algorithm that allows users to specify their prior knowledge in the form of simple queries in words or phrases and return query-related topics. It can then determine the probability of each article belonging to these topics. In this study, each policy area was treated as a topic and the GO Lab team provided keywords (Appendix A) for these policy areas so that the probability of articles belonging to them could be determined. As an example, the keywords for the category "homelessness" are as follows: 'housing', 'sleeping', 'accomodation', 'begging', 'sleep', 'residential', 'shelters', 'bed', 'streets', 'street', 'homelessness', 'refuge', 'shelter'. Three members of the GO Lab team were then asked to determine a probability threshold for each policy area. To determine the probability thresholds, the probability results of the articles were checked by the GO Lab Team to see if there was a clear threshold that would make a cut between articles classified as belonging to the correct policy sector and those with incorrect labels. Articles with a probability greater than the final threshold were assigned to the corresponding policy area. A total of 1,048 articles were classified as belonging to a policy area from the whole dataset that includes 1,952 articles.

### 4.6 Named Entity Recognition

In addition to the above information, named entity recognition techniques (as introduced in Section 3) - were also applied to extract named entities such as geographical locations (e.g. countries, cities, states, etc), organisations (e.g. companies, agencies, institutions, etc.), and relevant laws and regulations from the texts (these entities are tagged respectively as GPE, ORG and LAW in the models). Three named entity recognition models, "en_core_web_lg" and "en_core_web_trf" models from the python package 'spaCy' and the "ner-ontonotes-large" English model from 'Flair' were applied. "en_core_web_trf" is based on the RoBERTa-base transformer model (Liu et al., 2019). 'Flair' is a bi-LSTM character based model. All models were trained on the "OntoNotes 5" data source (Marcus et al., 2011) and are able to identify geographical locations, organisation names, and laws and regulations.

---

[7] https://datahub.io/core/country-codes

Although all models were able to identify the three types of entities, they performed differently. Misidentified entities were observed for each model. To reduce the false positive rate, an ensemble method was adopted, considering the entities that appear simultaneously in the results of any two models as the correct entities. Additionally, entities that were mentioned only once in the text and entities that contain words referring to journals and publications (e.g. 'Press', 'Journal', 'Publ', etc.) were removed, since they are less informative. Entities that were identified in more than one category were also removed, for example, entities that were classified as legal and organisational by different models, as these entities are ambiguous.

In relation to GPE entities, we do not consider hierarchical relationships between entities, nor do we differentiate between different entities with the same name. For example "Paris, France" is not distinguished from "Paris, Texas". It is also worth noting that pre-processing of articles is not equally effective for all files. For example there are cases where author information has been processed as part of the text of the articles. This is especially relevant for GPE entities, as author affiliations may be detected as entities mentioned in the text.

The GO Lab team examined the results and identified a list of noisy words that should not be included in the organisational entities in order to improve the quality of the entities. Post-processing methods were then applied to merge duplicate entities, e.g., "world bank", "the world bank", and "World Bank" into "World Bank". Some entities are in abbreviated form, e.g., "WB" instead of "World Bank" and abbreviated entities were replaced with their full names. To determine the full names of these entities, two rules were followed: 1) If the first character of each of the several words preceding an abbreviated entity can constitute that abbreviated entity, then they are considered to be the full name of that entity, 2) If the first word following a full-name entity is formed by the first character of each word in that entity, then that word is considered to be an abbreviation for the full-name entity.

**4.7 Semantic text similarity**

To support the exploration of the connections between different articles, the semantic text similarity technique introduced in Section 3 was applied. To calculate the similarity score between articles, the 10,000 most frequently mentioned words were first extracted from across all articles' titles and abstracts and the text vectorization technique TF*IDF was applied to convert each article's abstract into a feature vector based on these words.Using these TF*IDF feature vectors, the cosine similarity between different articles was calculated. Cosine similarity is a measure of the distance between two non-zero vectors. It ranges from -1 to 1. The closer the value is to 1, the more similar the two vectors are. It has been widely adopted in modern Information retrieval systems (Singhal, 2001).

## 5. SyROCCo Prototype Interactive Tool for Data Visualisation

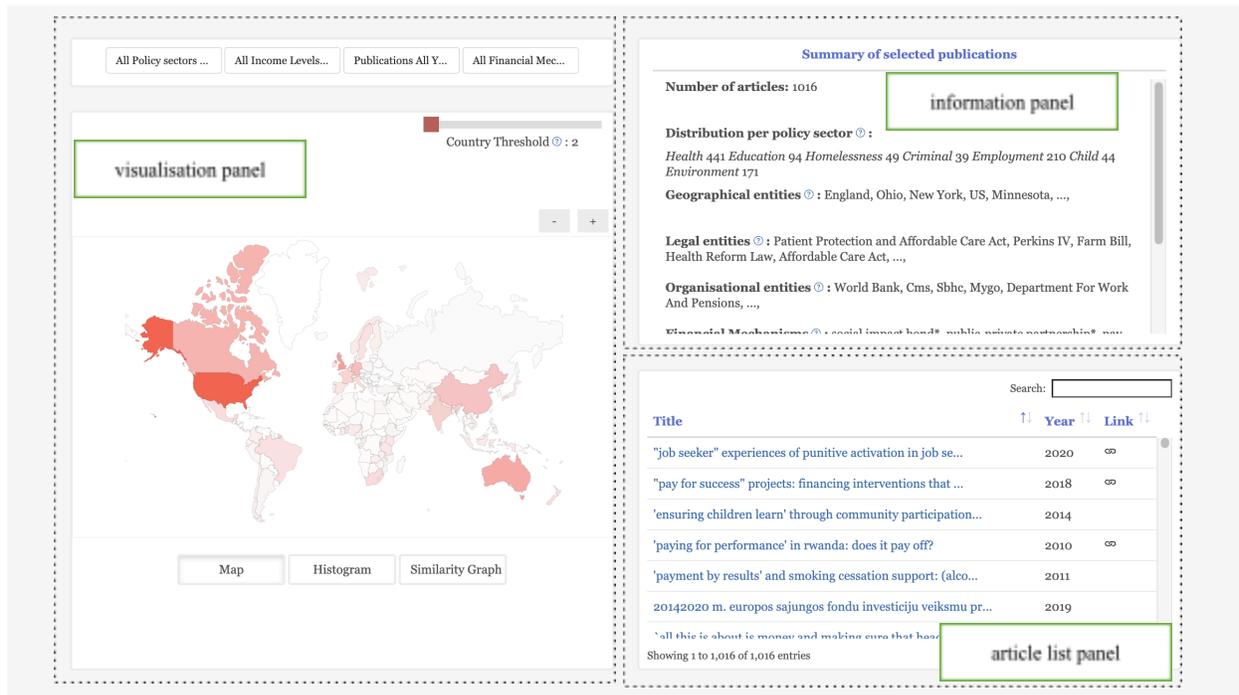

Figure 2: SyROCCo has three panels: visualisation panel (left), information panel (top right), and article list panel (bottom right).

We constructed the SyROCCo dataset by pre-processing the 1,952 studies of OBCs and extracting different types of information as described in section 4. The details of the dataset is presented in Appendix B. All data extracted using the machine learning techniques were collated in a visualisation tool that is publicly available.[8] Data dashboards were developed to support the exploration of the information extracted. The data dashboards are composed of three different panels: a visualisation panel, an information panel, and an article list panel. The visualisation panel provides different ways to visualise the data. The information panel displays information about articles that meet the selected criteria. The article list panel lists all articles that meet the selected criteria and allows users to easily search and access specific articles. Users can filter articles by policy areas, by mentioned countries' income level, by publication year, or by financial mechanisms (i.e. type of outcomes-based instrument).

---

[8] https://golab.bsg.ox.ac.uk/knowledge-bank/indigo/syrocco-ml-tool/

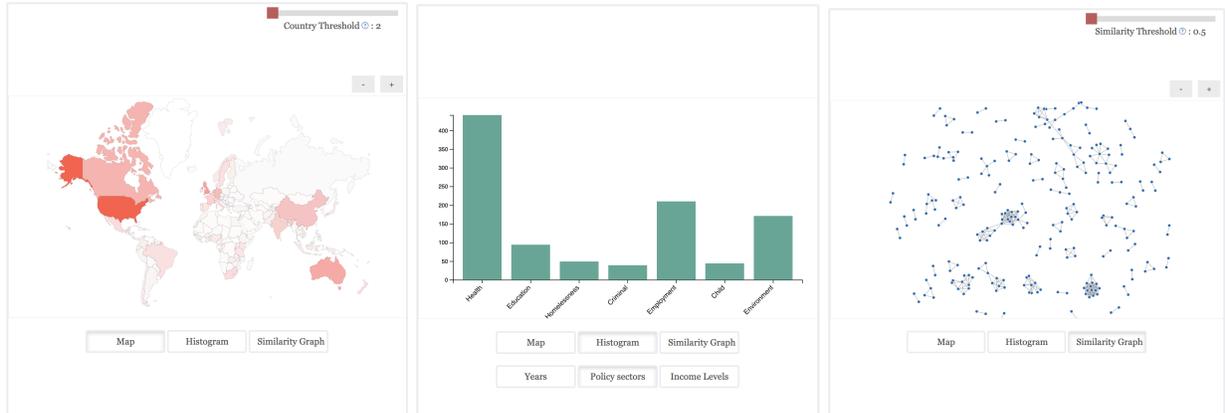

Figure 3: Three different types of interactive visualisations. World map (left). Histograms (centre), and Force-Directed Graph (right).

Three types of interactive visualisation plots are provided in the visualisation panel: a world map, histograms, and a force-directed graph. The interactive world map shows the distribution of articles that mention different countries. The darker the colour of a country on the map, the more articles mention it. By clicking on a particular country, users are able to access the information of all the articles mentioning the country. An overview of the information extracted from those articles is displayed in the information panel of the tool. This allows researchers and policymakers to access global information on outcomes-based contracting. To filter out less relevant articles, a slider bar is added to enable users to decide the minimum number of times a country must be mentioned in an article in order for it to be included in the summary.

For the interactive histograms, users are able to view the distribution of articles by published year, by corresponding policy areas, or by mentioned countries' income level. Clicking on a specific bar provides an overview of information for all articles under the specific criteria., e.g., all the articles published in 2020, all the articles belonging to the health policy area, or all the articles mentioning high income countries, etc.

The force-directed graph shows the semantic similarity between different articles. It allows researchers and policymakers to explore the connections between different articles. Each node in the graph represents an article. A solid line between two nodes indicates that the two corresponding articles are similar. The similarity score between articles was calculated using the method mentioned in section 3. If the cosine similarity between two articles is greater than a predefined similarity threshold, they are displayed in the graph and connected by a solid line. The similarity threshold is set by the "Similarity Threshold" slider bar shown on the top right side of the graph. By clicking on a particular node in the graph, users are able to access the detailed information of the corresponding article using the information panel, such as the information of its author, abstract, mentioned countries, mentioned INDIGO projects and so on.

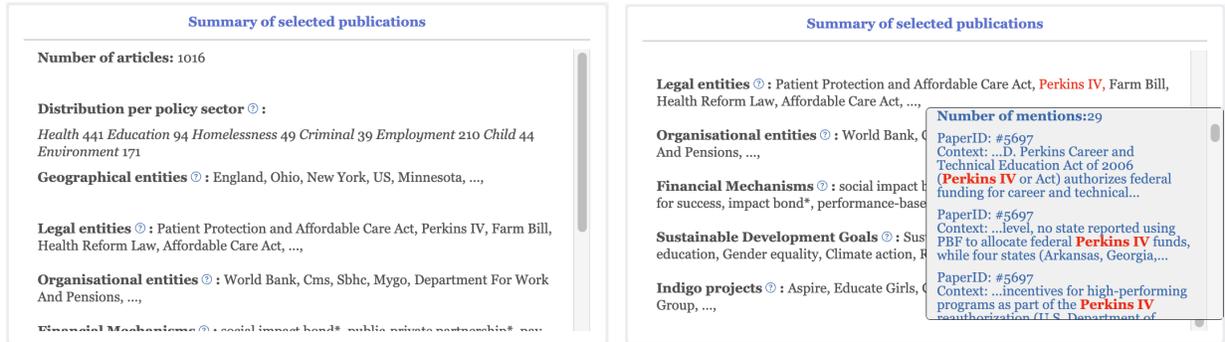

Figure 4: Information panel showing overview information about articles that meet a selected criteria.

The information panel is where users can view overviews of articles that meet a selected criteria or detailed information about a particular article. For the overview information, users can view the number of selected articles, the policy area distribution of these articles, geographical locations they mentioned, key organisations active in these articles, relevant laws and regulations, financial mechanisms mentioned, Sustainable Development Goals mentioned, and INDIGO impact bond projects mentioned. By clicking on an entity, users are able to view the context around the entity in the corresponding articles.

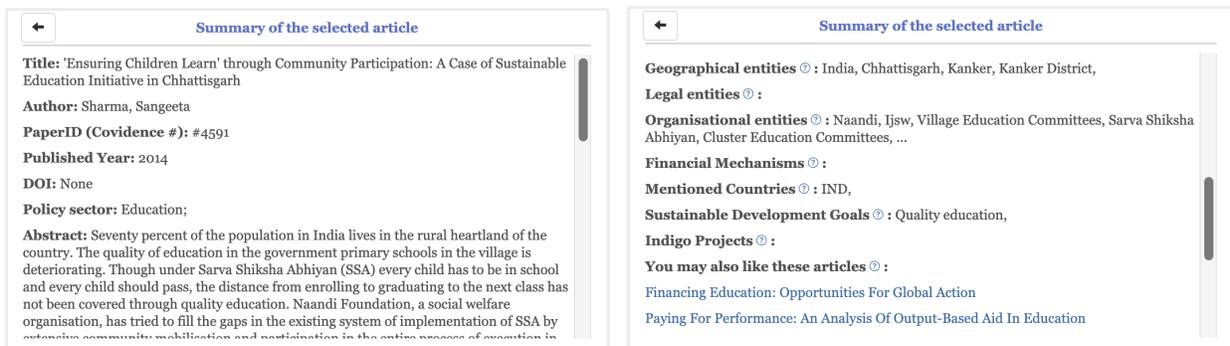

Figure 5: Information panel showing detailed information about a selected article.

When users select a specific article, they can also view its detailed information through the information panel. Users can view its title, author, published year, abstract, the policy area it covers, geographical locations mentioned, key organisations mentioned, relevant laws and regulations, financial mechanisms mentioned, countries mentioned, Sustainable Development Goals mentioned, and INDIGO projects mentioned. By clicking on an entity, users can view the context around the entity in the article. The 10 most similar articles to this one are also presented, thus enabling users to explore the links between them. By clicking on a similar article, users can view its details.

In summary, we sought to produce an intuitive dashboard that allows users to explore the data in our systematic review. For researchers, it aids the gathering of literature, reveals gaps in the evidence, and can inform refinement of research questions. For policymakers or practitioners, it provides timely access to evidence, can be tailored to relevant policy areas or countries, and may help identify pertinent legal frameworks or potential project partners. To further examine how far and how reliably the tool achieves these potential advantages, we conducted qualitative and quantitative evaluation.

## 6. Evaluation

**6.1 Qualitative evaluation**

The project followed an iterative, user participatory approach that integrated development and evaluation throughout. Hence, the team was collectively confident that the initial prototype embodied the requirements of GO Lab members. The next stage in development was therefore to solicit the views of a wider range of potential users. With this in mind, a demonstration of the initial prototype was made to the GO Lab 'Fellows of Practice', an international group of leading policy practitioners who collaborate with the GO Lab.

The Fellows of Practice agreed that being able to navigate articles with filters, i.e., policy sector, country, publication date etc., was useful:

> "I think that's going to be extremely useful, I'm sure we'll definitely use it, especially when you start researching something quite specific and thinking about the design of a programme being able to access all the literature around it."

However, the Fellows also felt that with long lists, going through the entire list could still be time consuming. It was suggested that being able to list the articles by their citation counts would help to identify the most important ones and thereby reduce the time and effort required.

Another comment was that a 'Netflix-style suggestion' of articles would be useful: "For example, if I pick one article, the tool should tell me 'You may also like this other paper'".

> "Like if I was looking for something on like in South America or something I want to focus on certain types of countries or interventions taking place in a certain region, which are the ones that are being shown through the filter are most likely to be, not because of my interest but more, you know that those seem to be clicked on."

Yet another comment was that it would be useful to be able to filter articles by type of outcomes-based instrument (social impact bond, results-based financing, etc.).

> "One question, would it be possible to get results by type of instrument, because you know within outcomes-based financing or however you call it, is lots of different things. And some of them would be more relevant sort of context and so just thinking from a practitioner's point of view, if I'm starting the design process for let's say an impact bond and then actually being able to filter by relevant articles for that type of instrument, that would be different from like a performance-based loan for example."

Finally, the Fellows remarked that the significance of the distance between clusters in the cluster map was not very clear.

"So, if I hover over it, I can get a sense of what the cluster is focused on, but then, if I go up here, I see this cluster and whatever it's talking about. There is a distance there with a couple other clusters. But is that distance and the relation between these clusters just random or is it meaningful that if there's a bunch of clusters in the Northwest quadrant it means something."

The review of the Fellows of Practice was valuable, and prompted us to introduce several suggested features in the tool: ordering the articles by citations, suggesting related articles, and filtering by types of instruments.

**6.2 Quantitative evaluation**

In addition to qualitative evaluation of the tool, we carried out a quantitative evaluation of the main machine learning techniques used in the tool for this dataset. It is important to note that the purpose of this project is not the evaluation of new machine learning methodologies *per se*, which has been done in the original publications and evaluations pertaining to the techniques (Reimers & Gurevych, 2019, Akbik et al., 2019, Liu et al., 2019, Fang et al., 2021). Rather, our evaluation offers important insights into the performance of these techniques in a complex field of application, namely a large corpus of heterogeneous social science literature on outcomes-based contracting.

We first evaluated the accuracy of policy area identification. For this assessment, we randomly selected a sample of 60 articles classified by the system under one of the seven categories employed (health, education, homelessness, criminal justice, employment and training, child and family welfare, environment and agriculture). We then asked a group of 5 subject matter experts to classify each of the articles under one of the above categories. The accuracy of random categorisation in this case is 14.3%. When comparing the results with those produced by our system, we obtained an accuracy of 71.7%. Further evaluations of this technique can be found in the original publication of the method used (Fang et al., 2021). We also evaluated the precision and recall of policy categorisation per category. The results for the precision are the following: health_precision = 100%, education_precision = 100%, homelessness_precision = 75%, criminal_precision = 66.7%, employment_precision = 31.2%, child_precision = 100%, environment_precision = 75%. The results for the recall are the following: health_recall = 68.7%, education_recall = 85.7%, homelessness_recall = 75%, criminal_recall = 100%, employment_recall = 100%, child_recall = 62.5%, environment_recall = 90%.

The quantitative evaluation revealed a mixed performance of the policy area identification model. While the overall accuracy of 71.7% is significantly higher than random chance, the performance varies considerably across categories. Some categories, such as health and education, exhibit perfect precision, meaning that all articles classified under these categories were relevant. Other categories, such as employment, show significantly lower precision (31.2%), indicating a high rate of false positives. In terms of recall, some categories, such as child and family welfare and health categories show the lowest scores, suggesting that the model might be missing relevant articles in this area.

Next, we carried out an evaluation of the Named Entity Recognition techniques. As mentioned above, these focus on the identification of three types of categories: geographical locations (GPE), laws and regulations (LAW), and organisations (ORG). This evaluation, carried out on a set of 20 randomly sampled articles, focused on the information retrieval of the 5 most mentioned results in each case, which is what is offered in the public tool. This was assessed using two metrics: precision@5 and recall@5. The first metric is calculated as the number of correctly identified entities divided by the number of retrieved entities. The second metric is the number of correctly identified entities divided by the number of entities in the text. All those numbers are calculated with a maximum of 5 for all quantities. The results obtained were: GPE entity recognition precision@5 = 69.2%, recall@ = 52.2%; LAW entity recognition precision@5 = 0%, recall@5 = 0%; ORG entity recognition precision@5 = 33%, recall@ = 40%. Details of these methods and other types of evaluations can be found on the 'spaCy' package page used in our implementation.

The evaluation of the Named Entity Recognition component of the tool revealed a less satisfying performance. Although the tool is relatively effective in identifying geographical locations (GPE), with a precision of 69.2% and a recall of 52.2%, it was less effective in the identification of laws and regulations (LAW). This could be explained by the variability in how laws and regulations are referenced in the text, as well as by the scarcity of such entities within the dataset (half of the selected and hand-coded articles did not contain any LAW entities). While the tool performed slightly better in identifying organisations (ORG), a precision of 33% and a recall of 40% also shows significant room for improvement. Here, evaluators noted that the tool does not accurately distinguish specific relevant organisations from general categories of entities (e.g. health maintenance organisation). Thus, the absence of precise definitions for entity types creates ambiguity, which lowers the accuracy of the model. The relatively low precision also indicates a high rate of false positives, where the tool falsely identifies entities as belonging to the organisations category.

The quantitative evaluation therefore reveals a more nuanced picture of the machine learning tools' performance. The policy area identification model demonstrates promising accuracy. But it also highlights the inherent difficulties in precisely categorising complex social science literature - a challenge even for humans. Some policy areas are less clearly delineated than others, exhibiting overlapping themes or less distinct terminology. For example, tools may struggle to differentiate between articles focused on employment and those focused on criminal justice as both may discuss issues related to reintegration into society. Similarly, policy areas such as homelessness often intersect with other policy areas, such as health, employment and housing (Grennan et al., forthcoming). In addition, many outcomes-based contracts tend to work with personalised services for people with complex needs. This means that one project might be working with homelessness, health and criminal justice issues at the same time (Carter et al., 2024).

The varying degrees of success for named entity recognition, apart from signalling a need for improvement in the models, also reveals the challenge of ambiguity in certain entities. Geographical entities are fairly clearly defined and bounded. Legal entities and organisations, by comparison, have hazier boundaries, and imply more scope for interpretation and error. Legal entities, for example, might include not only specific legislation (e.g. 'Affordable Care Act'), but also more general categories of regulation (e.g. 'carbon emission cap-and-trade regulations'), the names of government policies (e.g. 'Medicare'), or multilateral agreements (e.g. 'Paris Declaration on Aid Effectiveness'), and so on. That recall and precision was poor on these entities is perhaps rooted in this ambiguity. Refinement of the tools is no doubt possible, but this suggests that the difficulty is not solely technical, but also consists in the nature of the category that we are training the machine to 'understand'.

Another important aspect to note is the discrepancy between the qualitative evaluation and the quantitative evaluation. The qualitative evaluation highlighted the perceived usefulness and potential of the tool from the perspective of policy makers. Users valued the tools' interface and functionality as intuitive and meeting the needs of the target audience. While the positive feedback is encouraging, when read in conjunction with the quantitative analysis, significant risks become apparent. Users may be unaware of the limitations of the tool and the potential to miss relevant studies. Users may inadvertently exclude important evidence from their analysis, undermining the rigour and comprehensiveness of systematic reviews or of policy decision making.

Again, our quantitative analysis should not be interpreted as an evaluation of the machine learning techniques themselves, which has been done elsewhere. However, our evaluation of the techniques in a complex field of knowledge underscores the importance of designing transparent and interactive tools, which connect machine-generated output with the underlying data and allow for the interrogation of the results by humans. Tools need to be designed in view of enhancing efficiency and breadth of analysis (e.g. by highlighting relevant patterns in the data), but only while upholding the essential role of human judgement and expertise in interpreting complex information.

## 7. Discussion

Systematic reviews have become an important tool for researchers and policymakers wishing to gain a comprehensive and rigorous understanding of a particular field. They are time-consuming (Nussbaumer-Streit et al., 2021) and labour-intensive, however, and researchers often have to define searches narrowly in order to limit the number of articles to that which is feasible for human analysis.

During the development process, the project team, which consisted of GO Lab members and experts in machine learning, was able to build the common ground in terms of domain and technical expertise that is essential if machine learning projects are to deliver outcomes that actually meet their users' needs (Slota, 2020; Wolf, 2020; Arana-Catania et al., 2022). Inevitably, this takes time and so requires a continuing commitment by all those involved to contribute to the project. This was sustained by the scheduling of weekly meetings where progress could be demonstrated and reviewed, and objectives for the next meeting were agreed. To those that object that this is, in itself, time consuming, this project demonstrates that it nevertheless can deliver meaningful results. What is needed is for users to be convinced that time invested in such projects will be rewarded by savings in time and effort in their future research.

It is important to highlight that our prototype does not address every aspect of the traditional systematic review method. Notably, it does not address methodological quality. In a standard systematic review, each paper is assessed against methodological standards and assigned a quality category or score (Liabo et al., 2017). Prior studies have successfully trained ML tools to categorise high-quality evidence (Abdelkader et al., 2021) or assess risk of bias (Arno et al., 2022). They were developed on clinical evidence, however, while our evidence-base is from the social sciences and includes a significantly more heterogeneous range of study designs.

We initially explored the application of machine learning to quality appraisal (QA), but found it to be unviable in the project's timeframe. The nuanced knowledge and interpretation that human researchers used to assess a paper's quality, particularly in the social sciences where qualitative designs are prevalent, proved difficult to reduce to specific words or linguistic features. Unlike quantitative trials, qualitative research lacks widely agreed standards of methodological reporting (Carroll and Booth, 2015), perhaps making it less amenable to natural language processing techniques. Nevertheless, we do not reject outright that accurate ML tools may be developed for at least some domains of qualitative or mixed-method QA.

The team thus focused on addressing more amenable tasks. The prototype tool developed in this project provides its users with a range of methods of interrogating a dataset of articles, and visualisation techniques help users quickly determine the relevance of the information extracted. The qualitative analysis suggests significant enthusiasm, confirming the value of new approaches to navigating and making use of the rich data inherent in systematic reviews. The positive feedback from users in the qualitative evaluation suggests that the tool's interface and functionality are intuitive and go some way in meeting the needs of its target audience. In this way, this paper marks a promising step towards the development of machine learning support for identifying innovative forms of knowledge gathering from systematic reviewing processes. The qualitative evaluation conducted so far also suggests that the prototype will play a role in reducing the effort involved in the systematic review process. As such, the project has successfully delivered on its primary aim of developing a prototype tool to support the GO Lab's review of OBC and make the data accessible to our policymaker and practitioner audiences.

However, our study also highlights the challenges and potential serious risks of incorporating machine learning into systematic reviews. The quantitative evaluation indicates that the underlying machine learning models, while promising, require further refinement to improve accuracy and reliability, particularly in areas such as policy area identification and named entity recognition. The complexity of social science literature, where policy areas often intersect, methodology is diverse, and terminology can be ambiguous, poses challenges for machine learning techniques as well as human annotators. Although overall promising in its performance, these challenges became evident in the tools' difficulty of differentiating between policy areas with significant substantive overlap. Equally, the NER component struggled to accurately identify entities due to the inherent ambiguity of language in the text.

In particular, the discrepancy between the enthusiasm reflected in the qualitative evaluation of the tool and the more nuanced quantitative evaluation raises concerns about the potential for "automation complacency", where users may overestimate the tool's capabilities and overlook its limitations (Harbarth et al. 2024). These findings underscore the need for caution and transparency when incorporating machine learning into systematic reviews. While the qualitative evaluation indicates that our tool can be a valuable asset, it is important to acknowledge its current limitations and potential for error. Researchers and policy makers are encouraged to use our tool and similar machine learning-enabled tools as starting points for their analysis, but also to engage critically with the results, ensuring that human judgement remains central to the systematic review process. Further research is needed to refine techniques and develop strategies for integrating machine learning into systematic reviews in a way that upholds the rigour, transparency, and reproducibility that provide the bedrock to this methodology.

## 8. Conclusions and Future Work

The project team set out to develop a tool that would assist the GO Lab team in their description and analysis of a large corpus of articles on OBC, and the evaluation of the prototype shows that progress was made towards this goal. Following preprocessing, which improves the reliability of the subsequent results, four key tasks were facilitated by the machine learning tools. First, using text classification, 1,048 publications were categorised into a series of different policy areas from the total of 1,952 articles. Second, named entity recognition was used to extract organisations, laws, and geographical information associated with each publication. These first two tasks support the 'evidence mapping' stage in a systematic review - that is, describing the 'extent and nature of research that has been undertaken on an issue' (Gough et al., 2017, p. 125). Without the aid of machine learning, all such information would have to be extracted manually by reviewers - something which is too time- and resource-intensive to accomplish on nearly 2,000 articles. Machine learning offers evidence mapping capacities at a scale not otherwise practically achievable. Patterns, trends, and gaps in the evidence can be identified across a far greater number of publications.

The third task was to connect the systematic review database with the GO Lab's existing INDIGO database. Doing so augments the value of both resources and improves the coherence of the GO Lab's data offering, making it easier for users to navigate up-to-date information and publications about their projects of interest. Finally, the use of semantic text similarity presents reviewers with subgroups of publications that may share thematic content. This last task is more exploratory and interpretive. The machine learning tool produces the subgroups - thereafter it is up to humans to investigate what might be connecting them. Unlike the above example of evidence mapping, a specific step in the systematic review method is not directly accomplished by the semantic similarity tool. Nevertheless, exploring the 'similarity threshold' and the subgroups of connected articles may reveal certain thematic connections that a researcher would not have identified otherwise. It may therefore facilitate inductive analysis of the evidence base. This potential is yet to be explored further by the research team.

The results of all four tasks were visualised in an interactive web-based tool, freely available for use by researchers, policymakers, and practitioners. Users of the tool can now explore and navigate a large body of evidence that would otherwise have remained fragmented and dispersed. We have also publicly released both the dataset and the machine learning results produced during the project, thus aligning SyROCCo with principles of transparency, data-sharing, and collaboration.

From the perspective of project methodology, the prototype provides further evidence of how the effective development and deployment of machine learning tools requires following an interdisciplinary and participatory approach throughout, to foster an understanding among users of machine learning's capabilities and limitations, and to help data scientists to create context-sensitive, targeted, effective, trustworthy and ethical tools (Slota, 2020; Arana-Catania et al., 2022).

It is also clear that there will be opportunities to further develop and refine the prototype. Several next steps suggest themselves. First, we will explore the potential of machine learning tools for text summarisation, such that SyROCCo will not only extract key information, but also synthesise content from across multiple publications. Such evidence synthesis is a key part of the systematic review method. Second, it is important that our prototype is kept updated with the latest publications on OBC. As new reports are frequently published, we hope to be able to develop a pipeline tool that can screen relevant articles, extract their information, and add their details to the SyROCCo webtool. Finally, we want to make our tools as transparent as possible. We want to enhance the transparency and trustworthiness of SyROCCo by providing more comprehensive guidance and explanations as to how our tools function, their capacities and limitations. Doing so will also support reproducibility of the project.

The latest iterations of generative large language models (LLMs), such as OpenAI's GPT series or Google's PaLM, designed for prompt interaction, have garnered significant attention. While promising, those LLMs are designed for general prompt interaction (e.g., as a chatbot) and not to solve very specialised tasks (such as recognising named entities), and thus they offer even less control and 'human-in-the-loop' engagement, an essential component for examining the validity of results that is required for rigorous and trustworthy academic systematic reviewing. While generative prompt LLMs are powerful and can generate coherent and contextually relevant text, they operate as black boxes and can be unpredictable in their outputs. They do not provide the same deterministic and fine-grained control that traditional information retrieval or extraction techniques (like TF*IDF or LLMs trained specifically on named entity recognition) offer. Nevertheless, they do present intriguing possibilities. We are actively investigating the potential of leveraging generative prompt LLMs in different facets of the systematic review process, particularly in their capability to summarise content effectively.

Throughout these next steps, our aim is not to displace human effort altogether by automating the systematic review, but to build tools that enable human effort and expertise to be applied where it is of greatest value within the overall process. Achieving this will require careful attention to how to combine machine learning and human expertise within a 'human-in-the-loop' framework and not least to how to promote trust in machine learning tools (O'Connor et al., 2019). We also acknowledge that the protocols followed in systematic reviews can vary significantly. This suggests that attempting to create a single tool that can satisfy each and every variant is very likely to fail. Instead, what is needed is a toolbox of machine learning tools that can be flexibly assembled and configured to meet the objectives of specific reviews and their users.

Finally, having reflected on what the team as a whole has learned from this project, it seems to us likely that systematic review working practices will change to take advantage of the benefits that machine learning is capable of delivering. Exactly how they will change is unclear, but some possibilities suggest themselves. For example, if machine learning delivers on its promise of making the process significantly faster and with significantly less human effort involved, then this may offer a more direct role for policymakers who currently outsource this work to specialised units. Conceivably, this might even lead to policymakers increasingly taking on the task of systematic reviewing themselves. More realistic, perhaps, is the prospect of systematic reviewing experts and policymakers working more closely together as the process is conducted, enabling, for example, a more iterative evaluation of the outcomes and even a refinement of the policy questions to which the policymakers would like to have answers.

## Acknowledgements

We thank these funders for their contribution to this work: Department for Digital, Culture Media & Sports (grant ref: A2683); Children's Investment Fund Foundation (grant ref: 2104-06351); UK Research and Innovation (grant ref: MR/T040890/1); the UK Foreign, Commonwealth and Development Office (grant ref: 300539); UBS Optimus Foundation (grant ref: 51962); and the John Fell Fund (grant ref: 0012257).

## Funding statement

This research was supported by grants from the Department for Digital, Culture Media & Sports (grant ref: A2683); Children's Investment Fund Foundation (grant ref: 2104-06351); UK Research and Innovation (grant ref: MR/T040890/1); the UK Foreign, Commonwealth and Development Office (grant ref: 300539); UBS Optimus Foundation (grant ref: 51962); and the John Fell Fund (grant ref: 0012257). The funder had no role in study design, data collection and analysis, decision to publish, or preparation of the manuscript.

## Competing interests

None.

## Data availability statement

The data and code that support the findings of this study are available from: to be included upon acceptance of the article.

## Author contributions.

Zheng Fang: Conceptualization, Methodology, Software, Validation, Investigation, Writing - Original Draft, Visualization. Miguel Arana-Catania: Conceptualization, Methodology, Software, Validation, Investigation, Writing - Original Draft, Visualization, Supervision, Funding acquisition. Felix-Anselm van Lier: Conceptualization, Methodology, Software, Validation, Investigation, Resources, Data Curation, Writing - Original Draft, Visualization, Supervision, Funding acquisition. Juliana Outes Velarde: Conceptualization, Methodology, Software, Validation, Investigation, Resources, Data Curation, Writing - Original Draft, Visualization, Funding acquisition. Harry Bregazzi: Conceptualization, Methodology, Software, Validation, Investigation, Resources, Data Curation, Writing - Original Draft, Visualization, Funding acquisition. Mara Airoldi: Conceptualization, Methodology, Software, Validation, Investigation, Resources, Data Curation, Writing - Original Draft, Visualization, Supervision, Project administration, Funding acquisition. Eleanor Carter: Conceptualization, Methodology, Software, Validation, Investigation, Resources, Data Curation, Writing - Original Draft, Visualization, Supervision, Funding acquisition. Rob Procter: Conceptualization, Methodology, Software, Validation, Investigation, Writing - Original Draft, Visualization, Supervision, Project administration, Funding acquisition.

# Appendix A: Expert provided keywords for each policy area

**Health**
['diagnosed', 'physical', 'pneumonia', 'patient', 'illnesses', 'medications', 'disease', 'clinic', 'symptoms', 'physician', 'ailment', 'debilitating', 'treatment', 'treatments', 'medical', 'patients', 'doctor', 'treating', 'physicians', 'ailments', 'retardation', 'doctors', 'illness', 'psychological', 'hospitals', 'hospital', 'chronic', 'medication', 'mental', 'treat', 'suffering', 'psychiatric']

**Education**
['college', 'teaching', 'examination', 'kindergarten', 'taught', 'teacher', 'exams', 'numeracy', 'examinations', 'education', 'teachers', 'educational', 'students', 'school', 'grade', 'high', 'preschool', 'exam', 'graduate', 'pupils', 'elementary', 'grades', 'schools', 'attendance', 'literacy', 'student']

**Homelessness**
['asleep', 'housing', 'sleeping', 'accomodation', 'begging', 'sleep', 'residential', 'shelters', 'bed', 'streets', 'street', 'homelessness', 'refuge', 'shelter']

**Criminal justice**
['attorneys', 'rehab', 'ruling', 'imprisonment', 'lawyers', 'rehabilitation', 'recidivism', 'lawyer', 'convicted', 'attorney','supreme', 'judge', 'crimes', 'appeal', 'judges', 'charges', 'prosecutors', 'crime', 'prisoners', 'sentenced', 'imprisoned', 'jail', 'inmates', 'reoffending', 'prosecution', 'prison', 'court', 'murder','courts', 'trial', 'criminal', 'prisons', 'detention']

**Employment and training**
['work', 'jobseeker', 'labor', 'inflation', 'jobs', 'placement', 'wages', 'unemployment', 'careers', 'workplace', 'salaries', 'wage', 'internship', 'employment', 'joblessness', 'job', 'apprenticeship', 'jobless', 'labour', 'working', 'salary', 'career']

**Child and family welfare**
['family therapy', 'foster', 'foster care', 'neglect', 'domestic violence', 'domestic abuse', 'child abuse', 'adoption', 'family-centered practice', 'child protection' 'maltreatment', 'parental responsibility', 'fostering households', 'fostering agency', 'foster carer', 'fostering capacity', 'family support', 'unmatched children', 'out of home care']

**Agriculture and environment**
['Climate', 'Renewable energy', 'Carbon' 'decarbonisation', 'Net zero', 'Greenhouse gas', 'Emissions', 'Anthropogenic', 'Temperature rise' 'Agriculture', 'Farm', 'Land management', 'Land degradation', 'Biodiversity', 'Ecosystem', 'Ecology', 'Deforestation', 'Conservation', 'Green economy', 'Climate finance', 'Sustainability', 'Natural resources', 'Pollution', 'Resource degradation', 'Air quality', 'Waste', 'flooding']

# Appendix B: SyROCCo Dataset

This section describes the SyROCCo Dataset we constructed by pre-processing the 1,952 studies of OBCs and extracting different types of information mentioned in section 4. Each entry of the dataset contains the following information.

The basic information of each document is its title, abstract, authors, published years, DOI and Article ID. This information is screened by GO Lab team members during the screening phase of the project:
- Title: Title of the document.
- Abstract: Text of the abstract.
- Authors: Authors of a study.
- Published Years: Published Years of a study.
- DOI: DOI link of a study.

The probability of a study belonging to each policy area. We can determine the policy area of the study using the probability according to section 4.5:
- policy_sector_health: The probability of a study belongs to the policy sector "health".
- policy_sector_education: The probability of a study belongs to the policy sector "education".
- policy_sector_homelessness: The probability of a study belongs to the policy sector "homelessness".
- policy_sector_criminal: The probability of a study belongs to the policy sector "criminal"
- policy_sector_employment: The probability of a study belongs to the policy sector "employment"
- policy_sector_child: The probability of a study belongs to the policy sector "child".
- policy_sector_environment: The probability of a study belongs to the policy sector "environment".

Other types of information such as financial mechanisms, Sustainable Development Goals, and different types of named entities, according to section 4.4 and 4.6:
- financial_mechanisms: Financial mechanisms mentioned in a study.
- top_financial_mechanisms: The financial mechanisms mentioned in a study are listed in descending order according to the number of times they are mentioned, and include the corresponding context of the mentions.
- top_sgds: Sustainable Development Goals mentioned in a study are listed in descending order according to the number of times they are mentioned, and include the corresponding context of the mentions.
- top_countries: Country names mentioned in a study are listed in descending order according to the number of times they are mentioned, and include the corresponding context of the mentions. This entry is also used to determine the income level of the mentioned counties.

- top_Project: Indigo projects mentioned in a study are listed in descending order according to the number of times they are mentioned, and include the corresponding context of the mentions.
- top_GPE: Geographical locations mentioned in a study are listed in descending order according to the number of times they are mentioned, and include the corresponding context of the mentions.
- top_LAW: Relevant laws and regulations mentioned in a study are listed in descending order according to the number of times they are mentioned, and include the corresponding context of the mentions.
- top_ORG: Organisations mentioned in a study are listed in descending order according to the number of times they are mentioned, and include the corresponding context of the mentions.